\pdfoutput=1

\documentclass[11pt]{article}

\usepackage[preprint]{acl}

\usepackage{times}
\usepackage{latexsym}

\usepackage[T1]{fontenc}

\usepackage[utf8]{inputenc}

\usepackage{microtype}

\usepackage{inconsolata}

\usepackage{graphicx}

\definecolor{lightgray}{rgb}{0.97,0.97,0.97}

\usepackage{xcolor}

\usepackage{booktabs}
\usepackage{adjustbox}
\usepackage{multirow}
\usepackage{arydshln}  
\usepackage{colortbl}  
\usepackage{listings}
\usepackage{amssymb}
\usepackage{pifont}%

\newcommand{\cmark}{\ding{51}}%
\newcommand{\xmark}{\ding{55}}%

\title{PersonaBench: Evaluating AI Models on Understanding Personal Information through Accessing (Synthetic) Private User Data}

\author{Juntao Tan, Liangwei Yang, Zuxin Liu, Zhiwei Liu, Rithesh Murthy\\ 
{\bf Tulika Manoj Awalgaonkar, Jianguo Zhang, Weiran Yao, Ming Zhu, Shirley Kokane}\\ 
{\bf Silvio Savarese, Huan Wang, Caiming Xiong, Shelby Heinecke} \\
        Salesforce AI Research, USA}

\begin{document}
\maketitle
\begin{abstract}



Personalization is critical in AI assistants, particularly in the context of private AI models that work with individual users. A key scenario in this domain involves enabling AI models to access and interpret a user’s private data (e.g., conversation history, user–AI interactions, app usage) to understand personal details such as biographical information, preferences, and social connections. However, due to the sensitive nature of such data, there are no publicly available datasets that allow us to assess an AI model’s ability to understand users through direct access to personal information.

To address this gap, we introduce a synthetic data generation pipeline that creates diverse, realistic user profiles and private documents simulating human activities. Leveraging this synthetic data, we present PersonaBench, a benchmark designed to evaluate AI models’ performance in understanding personal information derived from simulated private user data.

We evaluate Retrieval-Augmented Generation (RAG) pipelines using questions directly related to a user’s personal information, supported by the relevant private documents provided to the models. Our results reveal that current retrieval-augmented AI models struggle to answer private questions by extracting personal information from user documents, highlighting the need for improved methodologies to enhance personalization capabilities in AI. 



\end{abstract}

\section{Introduction}
\begin{figure*}[t]
    \centering
    \includegraphics[width=1.0\linewidth]{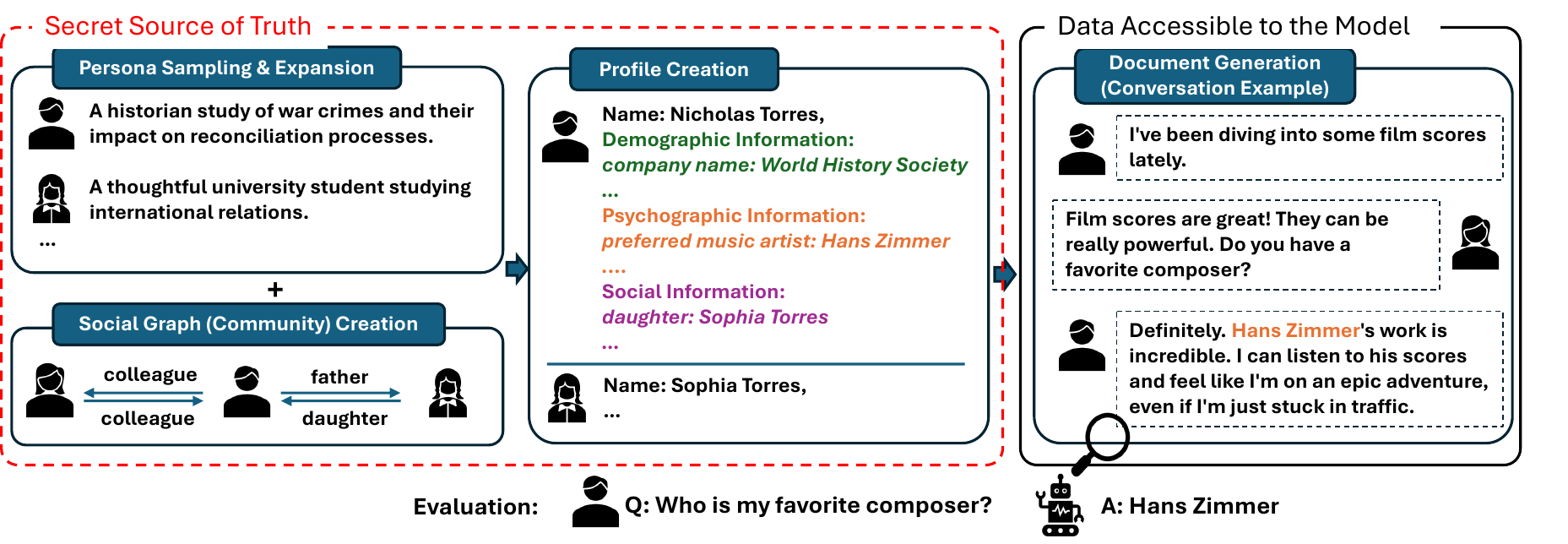}
    \caption{An illustration of the data generation pipeline and an example of a personal question used in the evaluation. The synthesized user profiles remain inaccessible to the models, which must rely solely on the user’s private documents to answer the questions. All examples are entirely synthetic and do not represent real people.}
    \label{fig:overview} 
\end{figure*}

LLM-based AI assistants are increasingly expected to provide personalized responses \cite{li2024personalized, salemi2023lamp, zhang2023memory, li2024personal}. This is particularly important for private AI models serving individual users. To achieve such personalization, it is crucial for the AI to understand key personal attributes, such as the user’s occupation, educational background, social connections, and preferences. For example, if a user asks the AI model for vacation recommendations, the model should consider the user’s preferred climate, travel budget, and previously visited destinations that they enjoyed.

However, this information is rarely provided explicitly to the model. Therefore, RAG has emerged as a popular solution in both research and commercial applications \cite{wang2024unims, salemi2024comparing, shi2024eragent}, where a retriever model takes the user’s query and identifies the most relevant information from available private user documents. These retrieved information are then combined with the original query and passed to the LLM for generating the final response.

Despite the potential of this approach, relying solely on RAG systems to ensure personalization may be an oversimplification. Retrieving and interpreting personal information is inherently complex and challenging. In practical settings, user data are often noisy, valuable personal details may be fragmented, and personal attributes can change over time. When deploying such systems, the true effectiveness of the pipeline remains unclear. This uncertainty comes largely from the lack of publicly available user documents paired with ground-truth personal information because of the sensitivity and privacy concerns of user data. Without publicly available standardized evaluation resources, it is difficult to objectively assess and improve these personalized AI assistants.

To address this gap, we introduce a synthetic data generation pipeline that produces realistic private user data. The pipeline begins by creating diverse and comprehensive user profiles as characters that include biographical details, personal preferences, and social connections. To enhance realism, these profiles are not generated in isolation, instead, they form interconnected social communities, ensuring that each connection aligns with the characters’ social attributes. These synthetic profiles then serve as the ground truth for generating various types of private user documents, such as conversation histories with socially connected individuals, interactions with AI models, and purchase histories. These private documents simulate realistic daily activities and naturally reveal their personal attributes. As a result, they can be used for standardized evaluations of AI models’ ability to extract relevant personal information for personalized response generation.

However, creating such datasets introduces several complexities: 1) It is difficult to ensure that the personal attributes within user profiles remain diverse, especially as the dataset grows. 2) Generating social relationships that feel natural and realistic while remaining consistent with individual personas requires a carefully designed strategy. 3) Synthesizing documents that faithfully reflect personal attributes while maintaining authenticity, such as incorporating necessary noise and time-sensitive information, poses significant challenges. 


To address the aforementioned challenges, the proposed data generation pipeline integrates persona sampling, social graph creation, multi-step profile completion, and multi-type private document generation. Each of these components is thoroughly explained in the following sections. We also note that while GPT-4o was used to synthesize all user profiles and documents in PersonaBench, the data generation pipeline is generally adaptable to other large language models.

In addition, to gain insights into an AI model’s ability to understand personal information by accessing private user data, we craft a diverse set of personal questions for each generated profile. Each question targets different attributes, and its answer can be found in their relevant document chunks. This setup allows us to assess models from multiple perspectives, ultimately creating a novel personalization benchmark. Figure \ref{fig:overview} illustrates the overall data generation and evaluation process. It is important to note that the generated user profiles are never directly shown to the evaluated models. Instead, the profiles only serve as the source of truth for producing private documents—documents that the AI models can fully access. This arrangement simulates a realistic scenario in which a private model must rely solely on available user documents to extract personal information.

Finally, we evaluated various RAG models equipped with retrievers of different sizes and state-of-the-art base LLMs. The results indicate that relying solely on the existing RAG pipeline is insufficient for effectively answering personal questions, underscoring the need for more advanced methodologies and systems to enhance personalization capabilities.


\section{Related Works}
\subsection{LLMs for Persona Modeling and Grounded Generation}
Our data generation pipeline relates to two major lines of existing research. The first focuses on using LLMs to generate persona-grounded data, predominantly in the form of dialogues. Early work by \cite{zhang2018personalizing} pioneered the concept of training small chit-chat models on dialogues grounded in personas composed of five simple sentences. Subsequent studies \cite{madotto2019personalizing,liu2020you} further improved the quality of such generated conversations. More recently, as LLMs have become increasingly prominent, researchers have explored using prompting techniques to produce high-quality dialogues grounded in given profile sentences \cite{lee2022personachatgen,jandaghi2023faithful}. While these efforts focus primarily on dialogue data, our approach extends this idea by generating multiple types of user documents based on comprehensive user profiles connected through social communities. Moreover, we produce a higher volume of utterances that occur over longer periods, making the data more realistic. Outside of dialogue generation, other work uses LLMs to simulate human behaviors, such as generating self-reports \cite{tavast2022language}, completing questionnaires \cite{hamalainen2023evaluating}, or simulate social interactions \cite{park2022social, park2023generative} .

The second line of research involves using LLMs to generate diverse personas directly. For example, \cite{chan2024scaling} introduces a method to produce large-scale persona sets by prompting LLMs to imagine who might have written a particular passage. In our work, we aim to create complex, information-rich user profiles. While our end goal differs, the publicly available personas from \cite{chan2024scaling} can serve as a resource to enhance the diversity of our generated user profiles.

\subsection{Evaluating LLMs on Personalized Generation}
Several recent benchmarks have started to evaluate the ability of LLMs to handle personalized generation tasks. For example, Tau-bench \cite{yao2024tau} uses LLMs to simulate characters and assesses an AI agent’s capability to gradually fulfill user requests throughout a conversation. In this setup, the user’s initial state is defined by a system prompt. Similarly, AppWorld Benchmark \cite{trivedi2024appworld} evaluates code agents tasked with completing user requests, using persona descriptions defined within various “task scenarios.” However, these benchmarks emphasize calling external APIs to fulfill personal requests while paying relatively little attention to the complexity of user profiles. Consequently, the tested scenarios rely on overly concise personal information that is straightforward to obtain, differing from our focus on accurately understanding more complex personal attributes rather than merely calling functions.

Another benchmark, LaMP \cite{salemi2023lamp}, specifically targets the evaluation of LLM personalization abilities. However, LaMP’s tasks—such as simulating a user’s writing style when composing emails or predicting whether a user would cite a given paper based on previous publications—differ from our primary goal. Their emphasis lies more in stylistic or behavioral imitation rather than the accurate extraction and interpretation of multifaceted personal information, placing it outside the main scope of our work.




\section{Stage 1: User Profile Synthesis}
The first stage of our data generation pipeline focuses on creating diverse, comprehensive user profiles, each representing a unique “character.” These synthetic individuals possess various attributes such as occupations, eating habits, and favored activities. In addition, they are socially connected and naturally forming different communities.

This stage requires careful planning, as the information in each user profile will serve as the authoritative source for generating all subsequent private documents. It will also guide the creation of personalized questions and answers. In the following sections, we describe each step of the user profile creation process, as well as the key considerations and solutions implemented at each stage.

\subsection{Profile Template Definition}
The first step is to define a profile template for each user, outlining the categories of personal attributes they should possess. We base our design on the hierarchical user template approach introduced in \cite{lee2022personachatgen}, which leverages social science research to determine the types of attributes a person may have. Building on this framework, we created a template for PersonaBench that organizes each user’s profile into three meta-categories:

\begin{itemize}
    \item Demographic Information: Basic details such as age, gender, occupation, and place of residence.
    \item Psychographic Information: Individual preferences across various topics, such as favored restaurants or hobbies.
    \item Social Information: Details regarding how the user interacts with and is connected to others within the community.
\end{itemize}

While every user profile includes these three meta-categories, the specific subcategories may differ between users. For example, one user’s profile might list “owned pets” as a subcategory, while another user’s profile may not include this information. These subcategories can be easily adjusted or extended to accommodate a variety of attributes. An example of the hierarchical structure of a user profile is provided in Appendix \ref{sec:profile_example}.

\subsection{Persona Sampling and Social Graph Creation}
When creating diverse characters by populating predefined profile templates, ensuring variety is crucial. A realistic group of individuals typically includes a wide range of backgrounds, statuses, and preferences. However, simply asking LLMs to fill empty profile templates often leads to significant duplication, especially as the number of generated users increases. This issue has been thoroughly discussed in \cite{chan2024scaling}. To address this challenge, they suggest including a brief, synthetic persona description alongside the query, which can sufficiently shift the LLM’s generation distribution. These short persona descriptions are typically one-sentence profiles (e.g., “A historian specializing in the study of war crimes and their impact on reconciliation processes.”).

Building on this idea, before generating a new community of characters, we randomly sample a set of persona descriptions from the public dataset released by \cite{chan2024scaling}, combining them with names generated using a random name generator. We then integrate these sampled personas into the profile generation prompt. This approach helps boost diversity and reduce duplication across the generated user profiles.

First, we create social information before filling other parts of each user’s profile because many personal attributes must be grounded in their social context. For example, colleagues should work at the same company. In this step, a group of individuals and their connections to one another are generated and represented as a social graph. We begin by randomly sampling $3$ short personas, then prompt the LLM to determine how these three individuals might be interconnected. Building on this initial trio, we use the LLM to expand the social graph by introducing additional individuals who are likely connected to the existing group. Each sampled or synthetic persona is integrated into the network, forming a social graph with relationships. After this initial generation, we perform post-processing on the graph’s edges to ensure that the relationships are symmetric, consistent, and error-free.

Figure \ref{fig:social} presents an example of a social graph from one community. Refer to Appendix \ref{sec:social_graph_details} for the sampled and synthetic persona descriptions of each node. Appendix \ref{sec:prompt_example} provides a simplified example to showcase the high-level structure of the prompt used for generating social graphs.

\subsection{Profile Completion}
After generating the social graph and incorporating social information into each profile, we fill the remaining attributes through the following steps:
\begin{enumerate}
    \item \textbf{Step 1: Socially Grounded Attribute Generation:} In this step, we focus on attributes that must be anchored in existing social relationships. We provide the LLM with both the social graph and each node’s persona description, enabling the model to generate attributes that are consistent with the defined relationships.
    \item \textbf{Step 2: Rest Profile Completion:} Next, for the attributes not directly related to the social graph, we individually generate the remaining details for each character. This process considers each user’s persona as well as the attributes already produced, ensuring internal consistency and a well-rounded profile.
\end{enumerate}


\section{Stage 2: Private Data Synthesis}
In the second stage, we generate synthetic private data to reflect the realistic behaviors and daily activities of each character. This data is produced in three document types: conversations, user–AI interactions, and user purchase histories.

\subsection{Document Types}
\begin{enumerate}
    \item \textbf{Conversation Data:} These documents simulate conversations between users. For each conversation session in one user, we select another individual who is directly connected to them in the social graph and generate a conversation session reflecting their relationship and context.

    \item \textbf{User-AI Interaction:} These documents capture direct chats between users and the AI assistant. Users may ask questions or engage in casual discussions. Over time, the AI can accumulate long-term personal information about the user through these interactions.
    
    \item \textbf{Purchase History} Based on each persona’s preferences, we synthesize purchase histories that reveal individual tastes and consumer behaviors. The format is inspired by the Amazon review dataset,\footnote{\url{https://cseweb.ucsd.edu/~jmcauley/datasets/amazon_v2/}} but for simplicity, we retain only the most relevant features, such as the item’s title, description, brand, and categories.
\end{enumerate}
Each document type contains only a portion of a person’s personal information. To comprehensively understand an individual, all documents must be combined. Each document includes a sequence of sessions with timestamps, simulating the passage of time in real-world user data. Sessions may be contextually linked, e.g., subsequent conversations can revisit previously discussed topics, and purchase histories are generated to avoid duplicating the same items.

Examples of each document type are provided in Appendix \ref{sec:document_example}.

\subsection{Generation Strategy}
In a realistic scenario, a user’s private data may not exclusively contain information indicative of their personal attributes. In fact, most user data may be “noise,” offering little insight into personal details. Therefore, when generating documents, we produce both persona-related data and noise data. Additionally, we incorporate real-world events into user conversations to enhance realism. Finally, we allow for the possibility that some personal information may be updated over time, further increasing the challenge of the dataset. We introduce each generation strategy below.

\subsubsection{Personal Data Generation}
To produce data that reveals personal attributes, we randomly select one attribute of the target individual and prompt the LLM to generate a session that subtly discloses this attribute. We encourage the LLM to avoid explicit, direct statements. Different types of documents employ different prompt templates. A simplified example demonstrating the overall prompt structure for generating conversations is provided in Appendix \ref{sec:prompt_example}. Notably, conversations may reference attributes of third parties within the user’s social network, making it possible for such attributes to be queried during evaluation.
\subsubsection{Noise Data Generation}
We also generate noise data that does not reveal personal information. Such data may include discussions about the weather, general questions posed to AI models, or purchasing everyday items. When generating noise, we instruct the LLM to avoid utterance that could reveal personal preferences, details, or traits of any participant. Instead, the focus should remain on general topics (e.g., ``Did you hear about the upcoming weather?''). We control the ratio of noise to personal data, with higher noise levels expected to reduce retrieval accuracy.
\subsubsection{Real-world News Integration}
To further increase realism, we integrate real-world news into the user conversations. We utilize an external tool\footnote{\url{https://newsapi.org/docs/client-libraries/python}} to retrieve publicly available news articles that align with the time window of the user’s private data. When generating each user conversation, there is a 20\% chance that a news event will be included in the prompt as contextual background. 
\subsubsection{Information updating}
A user’s preferences may change over time. For example, someone who once enjoyed romantic movies may lose interest after a significant life event. In the generated dataset, there is a small chance (less than 1\%) that previously mentioned preferences will be updated in a later conversation. This evolution of personal attributes further increases the complexity and challenge of the dataset.

\section{Experiments}
In this section, we first present the dataset statistics for the profiles, documents, and associated questions used in the testing set. Next, we describe the selected RAG models and outline the implementation details of the testing procedure. We then introduce the evaluation metrics employed for both the retriever-only and end-to-end evaluations. Finally, we discuss the main evaluation results and report findings from additional ablation studies.
\subsection{Statistics}
\begin{table*}[]
\centering
\begin{adjustbox}{width=1.0\linewidth}
\begin{tabular}{lllllllll}
\hline
\multirow{2}{*}{}   & \multicolumn{2}{l}{Noise 0} & \multicolumn{2}{l}{Noise 0.3} & \multicolumn{2}{l}{Noise 0.5} & \multicolumn{2}{l}{Noise 0.7} \\ \cline{2-9} 
                    & \# Session  & \# Utterance  & \# Session   & \# Utterance   & \# Session   & \# Utterance   & \# Session   & \# Utterance   \\ \midrule 
Conversation        & 1116        & 12901         & 1537         & 17914          & 2131         & 24756          & 3810         & 44810          \\
User-AI Interaction & 269         & 3187          & 401          & 4439           & 561          & 5749           & 1005         & 9955           \\
Purchase History    & 43          & -             & 97           & -              & 164          & -              & 373          & -              \\ \hline
\end{tabular}
\end{adjustbox}
\caption{Document statistics.}
\label{tab:document_statistics}
\end{table*}

\begin{table}[]
\centering
\begin{adjustbox}{width=0.6\linewidth}
\begin{tabular}{lcc}
\toprule
                  & number & multi-hop \\ \hline
Basic Info        & 269   &    \xmark       \\
Preference        & 186   &    \xmark       \\
Social            & 127   &    \cmark       \\ 
Total             & 582  &     -      \\ \bottomrule
\end{tabular}
\end{adjustbox}
\caption{Question statistics.}
\label{tab:question_statistics}
\end{table}
\subsubsection{Profile Statistics}
Following the described data generation pipeline, we produce $5$ communities and then randomly select $3$ users from each community with their documents for testing. In total, the test set includes $15$ characters, each with corresponding questions and ground-truth answers for testing. Each character is associated with up to $48$ categories of personal information. Although each category may contain multiple entries, none exceed $5$.

\subsubsection{Private Documents Statistics}
Private documents are generated under different noise levels, specifically at noise ratios of 0.0, 0.3, 0.5, and 0.7. Table \ref{tab:document_statistics} summarizes the total number of sessions and utterances for each document type in the test set.

When generating documents for each individual, the attributes that will be queried are guaranteed to appear, while other attributes may or may not be included. This design ensures variability and realism in the data.
\subsubsection{Personal Q\&A Statistics}
The personal Q\&A includes three types of questions:
\begin{itemize}
    \item Basic Information Questions: Related to demographic attributes.
    \item Preference Questions: Related to psychographic attributes.
    \item Social Questions: Related to social attributes.
\end{itemize}
Multiple question templates are predefined for each attribute category to ensure that the questions appear in diverse forms. For example, when inquiring about a person’s birthplace, the question might be “Which city was I born in?” or “What city is listed as my place of birth?”

For each user, demographic and psychographic attributes are downsampled before constructing questions, but all social attributes are queried. Table \ref{tab:question_statistics} shows the number of questions in each category.

Most questions are single-hop. However, the social category includes multi-hop questions. For example, determining “What is my sister’s favorite movie?” requires identifying the sister first and then finding her preferred movie.

\begin{table*}[]
\centering
\begin{adjustbox}{width=0.9\linewidth}
\begin{tabular}{lllllllll|ll}
\toprule
\multirow{2}{*}{Retriever} & \multicolumn{2}{l|}{Basic Information} & \multicolumn{4}{l|}{Preference} & \multicolumn{2}{l|}{Social} & \multicolumn{2}{l}{Overall} \\
                                   & Recall & NDCG & Recall (Easy) & NDCG (Easy) & Recall (Hard) & NDCG (Hard) & Recall & NDCG & Recall & NDCG \\ \midrule
all-MiniLM-L6-v2          & 0.235 & 0.163 & 0.369 & 0.333 & 0.278 & 0.265 & 0.252 & 0.186 & 0.236 & 0.186 \\
all-mpnet-base-v2         & 0.283      &  0.224     & 0.317      & 0.269      &  0.283     & 0.285      & 0.247      &  0.194     &  0.267     &  0.229     \\
bge-m3                    & \textbf{0.335}      & \textbf{0.252}      & \textbf{0.394}      & \textbf{0.385}      & \textbf{0.351}      & \textbf{0.357}      & \textbf{0.340}      & \textbf{0.263}      &  \textbf{0.325}     & \textbf{0.280}      \\
\bottomrule
\end{tabular}
\end{adjustbox}
\caption{Retrieval evaluation at $0.5$ noise ratio.}
\label{tab:retriever_results}
\end{table*}

\begin{table*}[]
\centering
\begin{adjustbox}{width=0.9\linewidth}
\begin{tabular}{lllllllll|ll}
\toprule
\multirow{2}{*}{Model + Retriever} & \multicolumn{2}{l|}{Basic Information} & \multicolumn{4}{l|}{Preference} & \multicolumn{2}{l|}{Social} & \multicolumn{2}{l}{Overall} \\
                                   & Recall & F1 & Recall (Easy) & F1 (Easy) & Recall (Hard) & F1 (Hard) & Recall & F1 & Recall & F1 \\ \midrule
\rowcolor[HTML]{F2F2F2} \multicolumn{11}{c}{\textbf{GPT-4o}} \\ 
Ground Truth Context                & \textit{0.362}          &  \textit{0.372}                          & \textit{0.538}       &  \textit{0.562}                       & \textit{0.424}       & \textit{0.453}                        & \textit{0.476}       & \textit{0.425}                        & \textit{0.444} & \textit{0.453}            \\ \hline
all-MiniLM-L6-v2          & 0.135          &  0.121         & 0.260       & 0.279         & 0.189       & 0.216        &  0.206     & 0.214         &  0.180     & 0.182       \\
all-mpnet-base-v2         & 0.154          & 0.153          & 0.170       & 0.185        & 0.168       & 0.208         & 0.206       & 0.214         & 0.172      & 0.183       \\
bge-m3                    & \underline{0.195}          & \underline{0.198}          &  0.303      & 0.302        & 0.238       & 0.253         & 0.198       & 0.213        & \underline{0.237}      & \underline{0.241}       \\
\hline   
\rowcolor[HTML]{F2F2F2} \multicolumn{11}{c}{\textbf{GPT-4o-mini}} \\ 
Ground Truth Context                & \textit{0.424}          &  \textit{0.454}                          & \textit{0.663}       & \textit{0.652}                        & \textit{0.523}       & \textit{0.559}                        & \textit{0.571}       &  \textit{0.540}                       & \textit{0.502} & \textit{0.521}            \\ \hline
all-MiniLM-L6-v2          & 0.143          & 0.132          & 0.297       & 0.301         & 0.212       & 0.241        & \underline{0.444}    &  0.355        & 0.214      & 0.208       \\
all-mpnet-base-v2         & 0.161          & 0.156          &  0.293      & 0.280         & 0.258       &  \underline{0.290}       & \textbf{0.492}      & \textbf{0.402}         & 0.229      & 0.224        \\
bge-m3                    &  \textbf{0.212}         & \textbf{0.221}          & \textbf{0.330}       & \textbf{0.331}        &  \textbf{0.290}      & \textbf{0.317}         & 0.437       & \underline{0.387}        & \textbf{0.277}      & \textbf{0.281}       \\
\hline   
\rowcolor[HTML]{F2F2F2} \multicolumn{11}{c}{\textbf{GPT-4}} \\ 
Ground Truth Context                & \textit{0.333}          & \textit{0.331}          & \textit{0.653}       & \textit{0.604}         & \textit{0.513}       & \textit{0.499}         & \textit{0.317}      & \textit{0.230}        & \textit{0.429}     & \textit{0.405}       \\ \hline
all-MiniLM-L6-v2          & 0.120          & 0.115          & 0.285       & 0.296         & 0.201      & 0.211         & 0.159      & 0.167         & 0.161      & 0.166      \\
all-mpnet-base-v2         &  0.123        & 0.115         &  0.246      & 0.217         & 0.218       & 0.209        & 0.111       & 0.129         & 0.163      & 0.153      \\
bge-m3                    & 0.176          & 0.178          & \underline{0.307}       & 0.307         &  \underline{0.275}      & 0.254        & 0.198       & 0.213         & 0.228     & 0.223      \\
\hline 
\rowcolor[HTML]{F2F2F2} \multicolumn{11}{c}{\textbf{GPT-3.5-turbo}} \\ 
Ground Truth Context                & \textit{0.374}          & \textit{0.382}          &  \textit{0.518}      &  \textit{0.542}        & \textit{0.472}      & \textit{0.479}         & \textit{0.667}       & \textit{0.690}         &  \textit{0.460}     & \textit{0.470}       \\ \hline
all-MiniLM-L6-v2          & 0.126         &  0.114         & \underline{0.307}      & \underline{0.314}         & 0.169       & 0.200         & 0.286       & 0.287         &  0.182     & 0.183       \\
all-mpnet-base-v2         & 0.142         & 0.140          &  0.247      & 0.237         & 0.192       & 0.223         & 0.206       & 0.180  
& 0.181      & 0.182       \\
bge-m3                    & 0.176          & 0.175         & 0.305       &  0.297       &  0.232     & 0.250         & 0.198       & 0.179         & 0.224      & 0.222      \\
\bottomrule
\end{tabular}
\end{adjustbox}
\caption{End-to-end evaluation at $0.5$ noise ratio.}
\label{tab:e2e_results}
\end{table*}

\begin{figure}[t]
    \centering
    \includegraphics[width=1.0\linewidth]{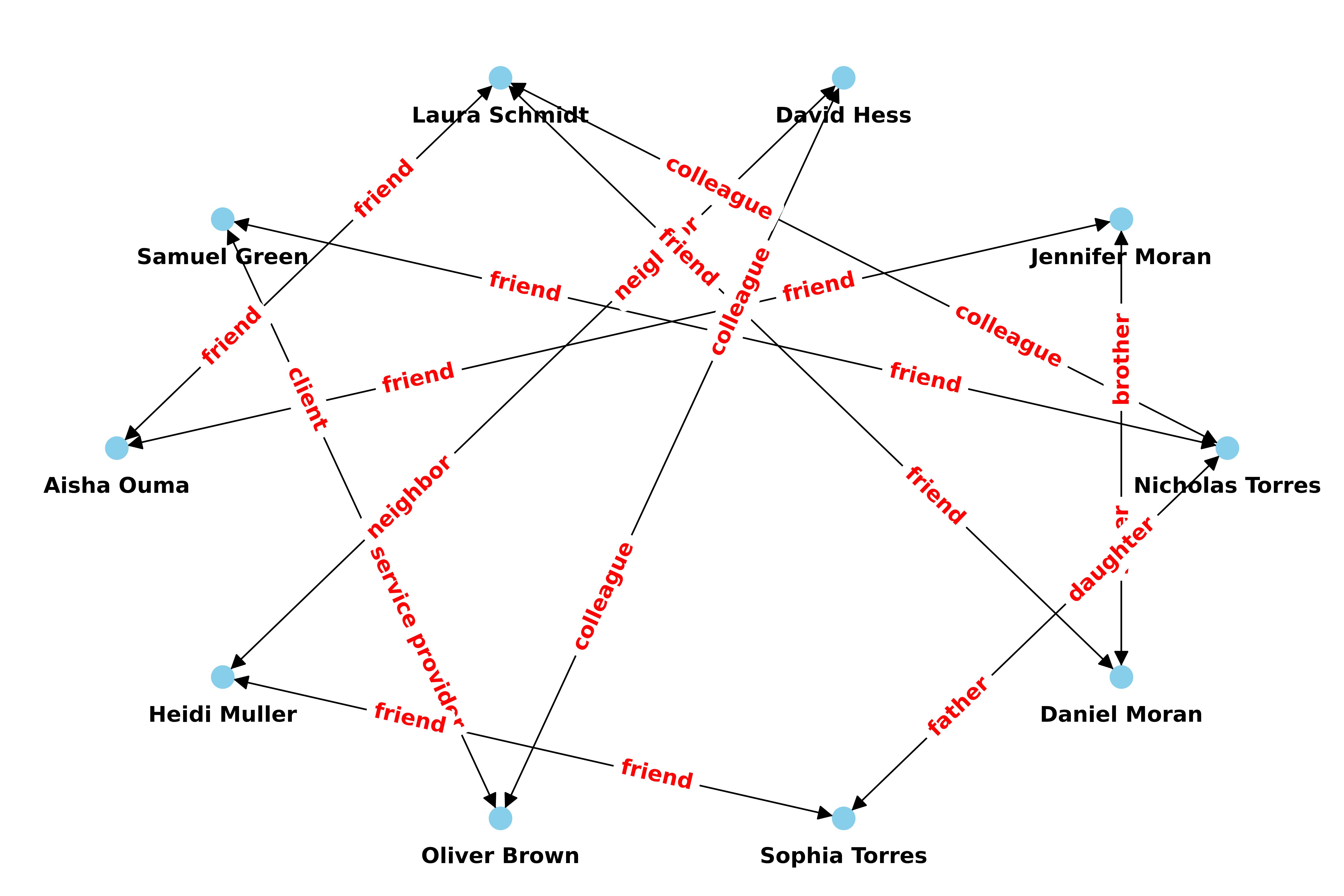}
    \caption{Social graph example. Refer to \ref{sec:social_graph_details} for the sampled and synthetic persona of each node.}
    \label{fig:social}
\end{figure}


\begin{figure}[t]
    \centering
    \includegraphics[width=1.0\linewidth]{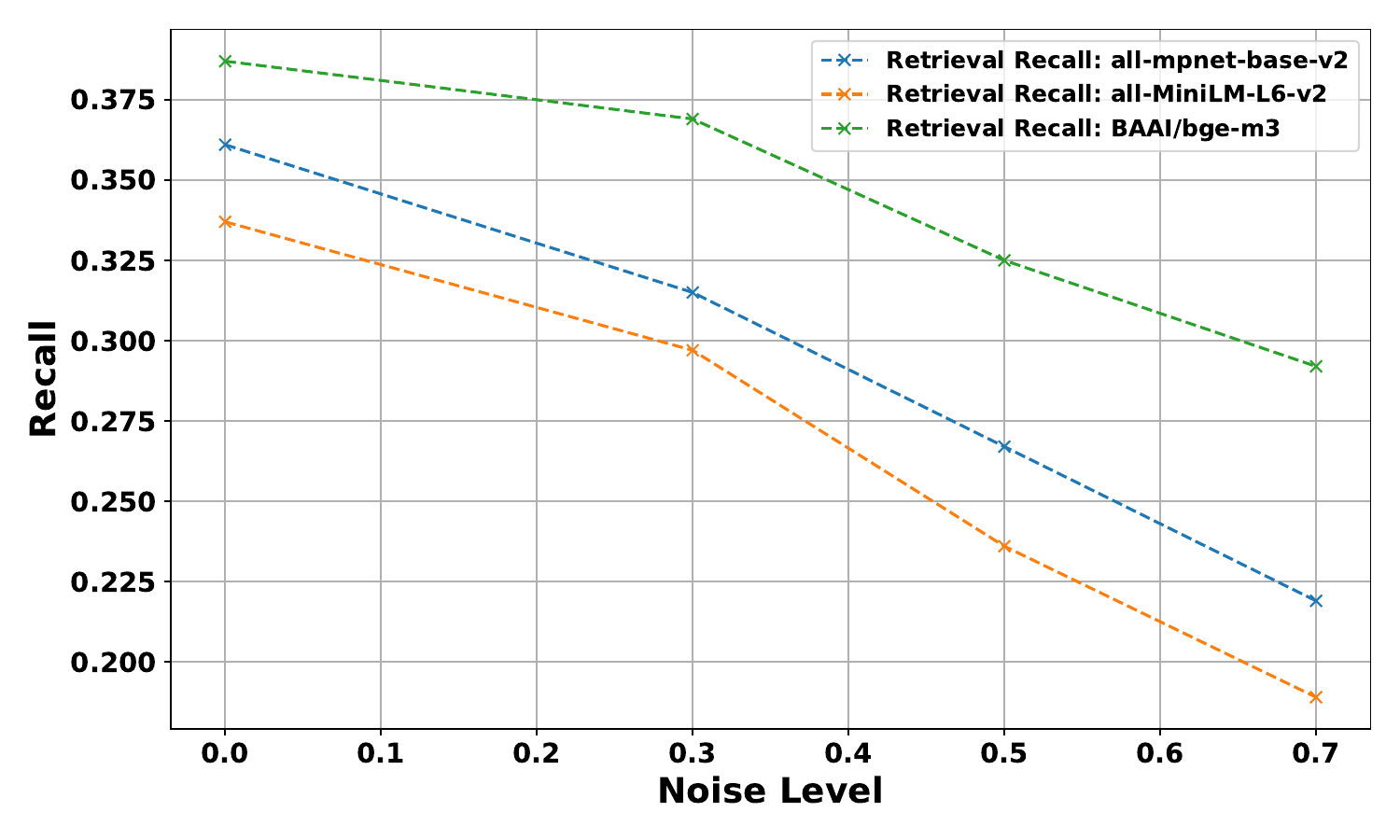}
    \caption{Retrieval performance vs. noise levels.}
    \label{fig:retrieval_noise}
\end{figure}

\begin{figure}[t]
    \centering
    \includegraphics[width=1.0\linewidth]{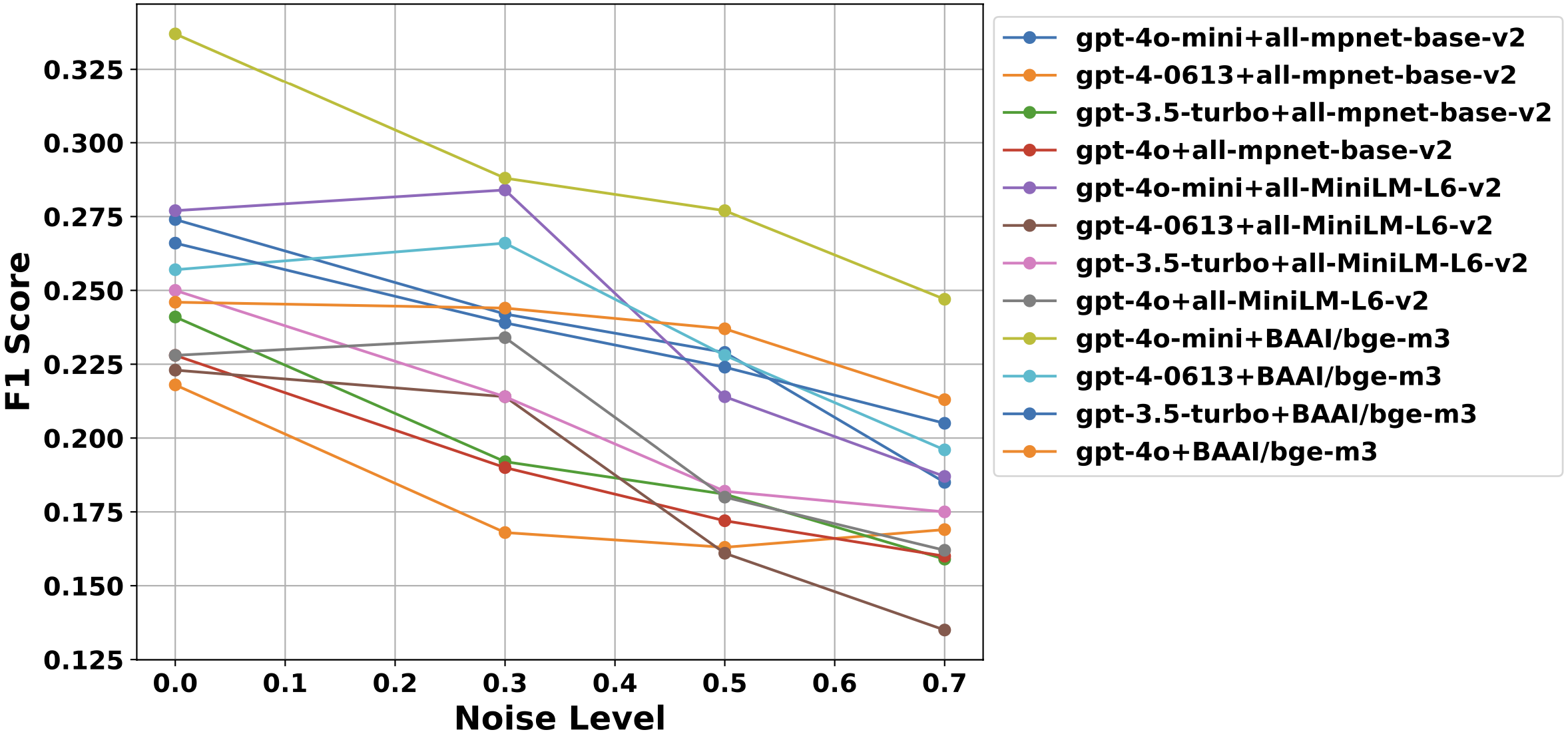}
    \caption{End-to-end performance vs. noise levels.}
    \label{fig:e2e_noise}
\end{figure}

\begin{figure*}[htbp]
    \centering
    \hspace{-30pt}
    \begin{minipage}[b]{0.38\linewidth}
        \centering
        \includegraphics[width=\linewidth]{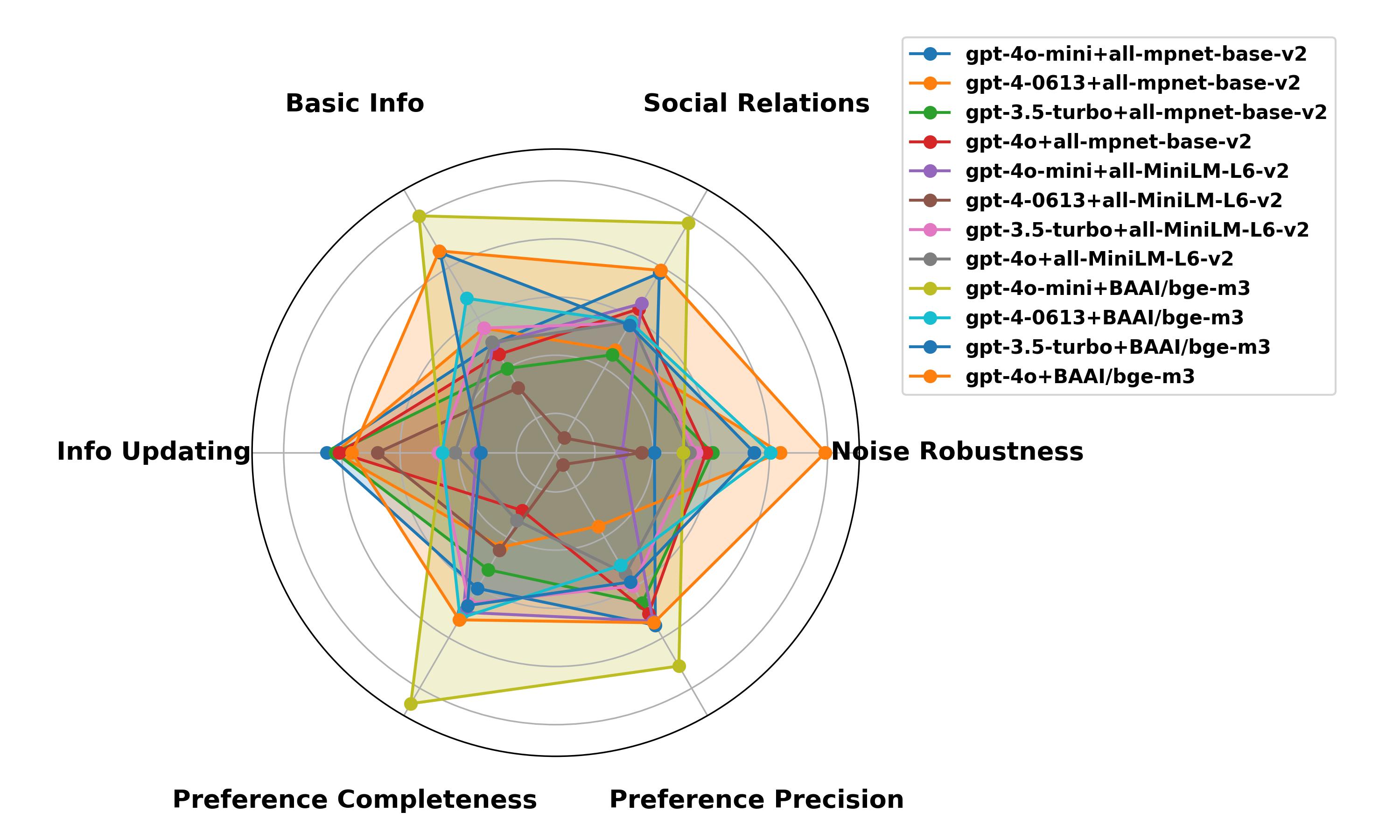}
        \caption{Performance comparison \\ on different dimensions (All models).}
        \label{fig:radar_all}
    \end{minipage}
    \hspace{-150pt}
    \hfill
    \begin{minipage}[b]{0.38\linewidth}
        \centering
        \includegraphics[width=\linewidth]{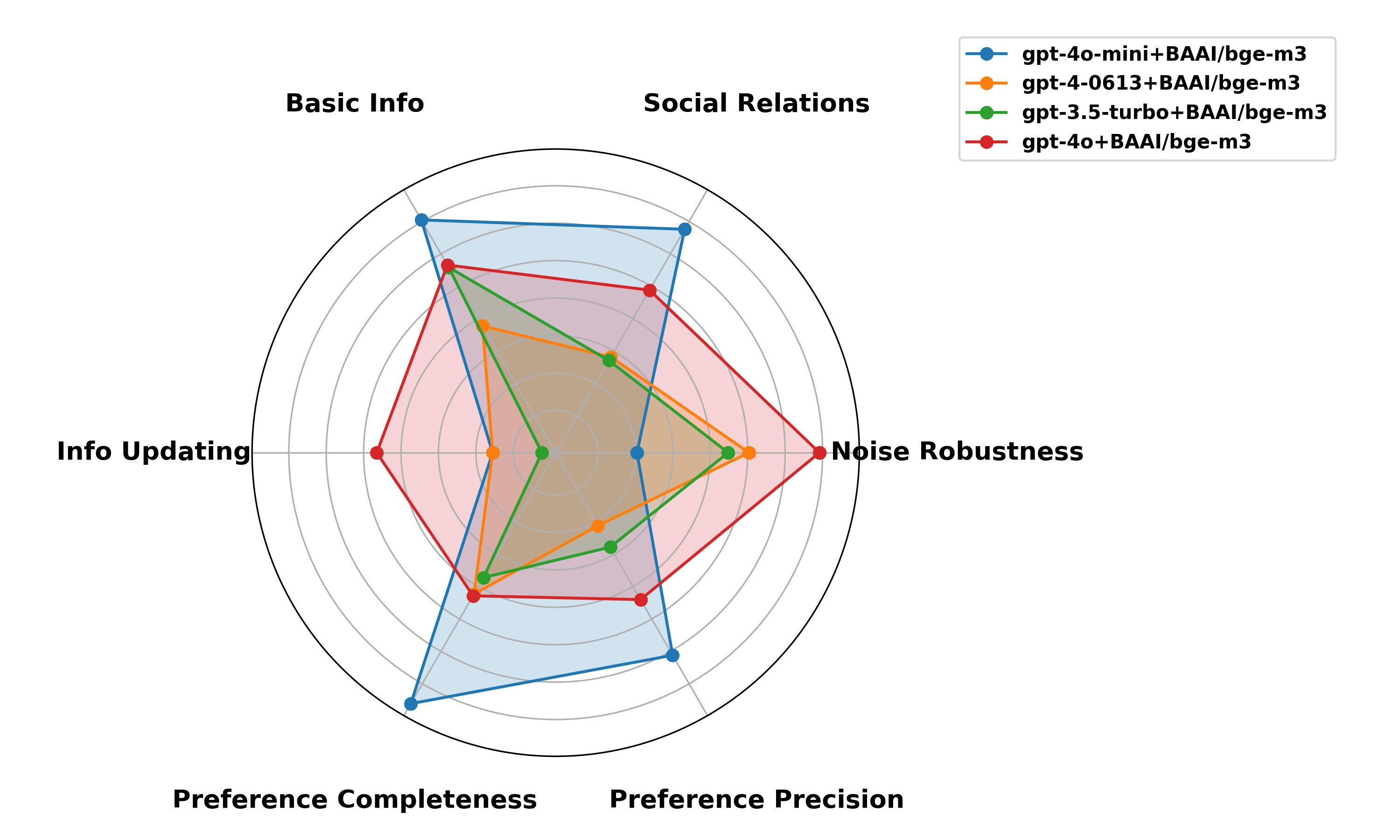}
        \caption{Performance comparison \\ on different dimensions (base models).}
        \label{fig:radar_LLM}
    \end{minipage}
        \hspace{-150pt}
    \hfill
    \begin{minipage}[b]{0.38\linewidth}
        \centering
        \includegraphics[width=\linewidth]{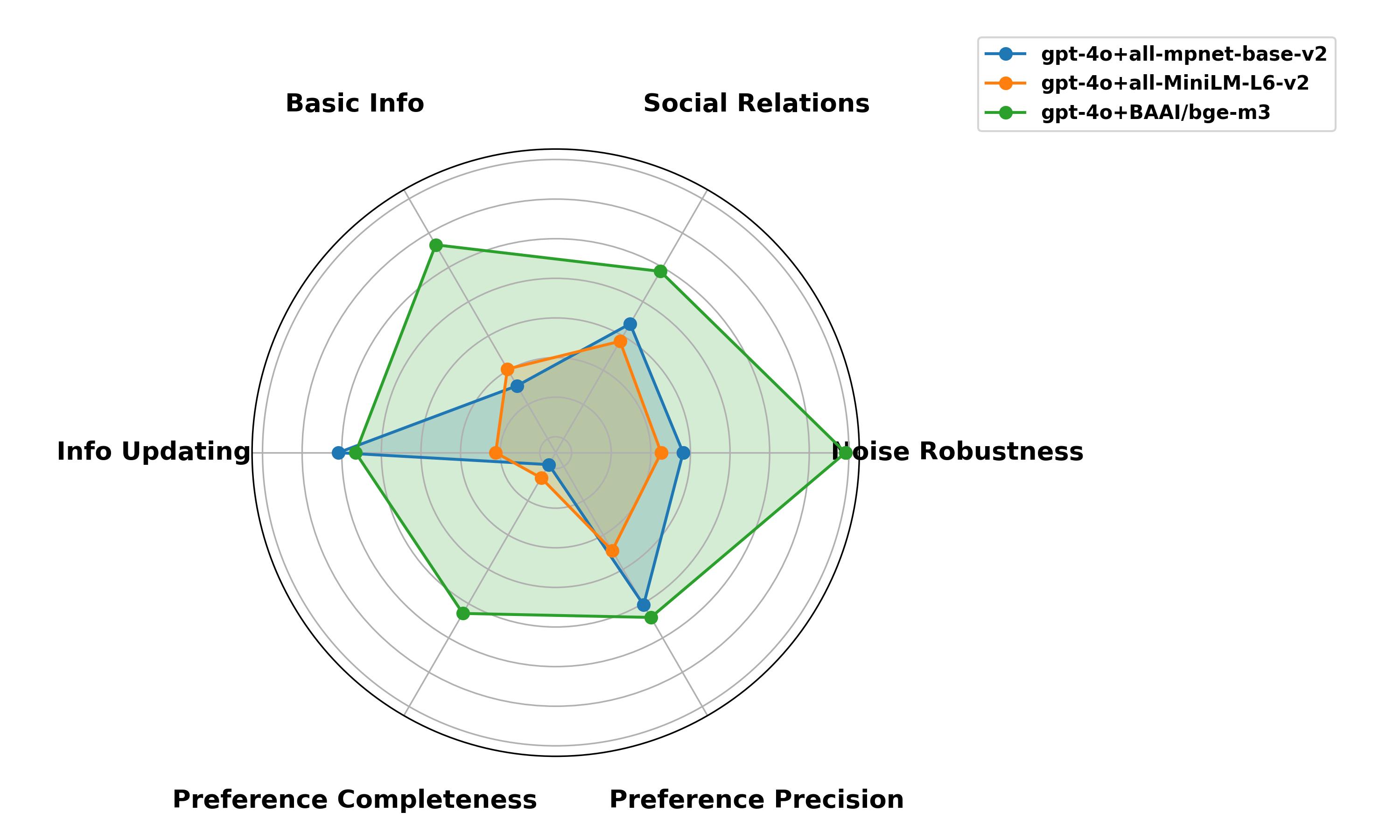}
        \caption{Performance comparison \\ on different dimensions (retriever models).}
        \label{fig:radar_retriever}
    \end{minipage}
    \hspace{-40pt}
\end{figure*}

\subsection{Model Selection and Implementation Details}
We evaluated a standard RAG pipeline, where a pretrained retriever model retrieves multiple sentence chunks that match the given query, and then concatenate them with the query as the LLM input.

For retriever models, we selected three dense retrievers from the SentenceTransformers library \cite{reimers-2019-sentence-bert}, each with different parameter sizes: all-MiniLM-L6-v2 (23M parameters)\footnote{\url{https://huggingface.co/sentence-transformers/all-MiniLM-L6-v2}}, all-mpnet-base-v2 (110M parameters)\footnote{\url{https://huggingface.co/sentence-transformers/all-mpnet-base-v2}}, and bge-m3 (567M parameters) \cite{chen2024bge}\footnote{\url{https://huggingface.co/BAAI/bge-m3}}.


For the base LLM, we tested four GPT variants: GPT-4o, GPT-4o-mini, GPT-4, and GPT-3.5-turbo. Combining these four LLMs with the three retrievers gives a total of 12 distinct RAG models.

We segment documents by session timestamps, treating each session as a natural chunk. The number of retrieved chunks is set to meet the maximum required for the most complex question in the dataset. For instance, if the most complex question requires five segments, we set the retrieval parameter to five for all questions.

In each session, we provide the timestamp, the actual content, and the involved person’s name to the model. Other details, such as the attributes used to generate the session, are omitted. This simulates what a real AI assistant would have access to when examining user documents.

\subsection{Evaluation Metrics}
We design two types of evaluations:
\begin{itemize}
    \item Retrieval evaluation: We assess only the retriever component of the RAG system, verifying whether the document sessions needed to answer the question are correctly retrieved. Recall and NDCG are used to measure retriever performance.
    \item End-to-end evaluation: We evaluate the entire RAG pipeline by testing whether the final answers to the personal questions are correct. As most ground-truth answers consist of multiple distinct terms, we use Recall and F1-score to measure how accurately the model’s predictions match the reference answers.
\end{itemize}

\subsection{Results}
We report both per-category and overall performance for retrieval and end-to-end evaluations. Within the preference category, we further divide questions into “easy” and “hard” based on whether the category contains fewer than five entries.

Table \ref{tab:retriever_results} presents the retrieval evaluation results at a $50\%$ noise level. The results indicate that larger retriever models generally achieve better performance. For example, the largest retriever model, bge-m3, outperforms all-mpnet-base-v2 and all-MiniLM-L6-v2 by $21.7\%$ and $37.7\%$ in Recall, and by $22.7\%$ and $50.5\%$ in NDCG, respectively. However, even the best retriever’s overall Recall is only $0.325$, indicating that more than half of the necessary information cannot be successfully extracted from irrelevant data. This underscores the challenge of our dataset for retrieval tasks.

Table \ref{tab:e2e_results} shows the end-to-end evaluation results. For each base model, combining it with bge-m3 shows the best performance, which is consistent with the retrieval evaluation. Comparing the base models, while GPT-4o is considered state-of-the-art, GPT-4o-mini surprisingly achieves the best overall results. Additionally, GPT-3.5-turbo produces comparable outcomes to GPT-4, suggesting that a generally superior model does not necessarily excel at personal Q\&A. We also report the performance when providing ground-truth context. Although this scenario significantly outperforms using retrieved context, the Recall scores are still around 50\%. This implies that some information is conveyed very implicitly, and the base models may not fully recognize it.

We conducted two ablation studies to gain deeper insights. First, we examined how retrievers and RAG models perform under varying noise levels. As shown in Figures \ref{fig:retrieval_noise} and \ref{fig:e2e_noise}, increasing noise leads to a consistent decline in retrieval performance, which in turn lowers overall RAG performance. This suggests that a model’s ability to extract personal information from noisy content is a key factor in its effectiveness.

Second, we evaluated RAG models along several dimensions, including Basic Information (F1 score on basic information questions), Social Relations (F1 score on social information questions), Noise Robustness (the difference in F1 scores between 0\% and 70\% noise levels), Preference Precision (precision on preference questions), Preference Completeness (recall on preference questions), and Information Updating (one minus the percentage of outdated information retrieved).

Figure \ref{fig:radar_all} visualizes performance across these dimensions, all the scores are normalized using z-score normalization. When holding the retriever model constant and comparing only the base LLMs (Figure \ref{fig:radar_LLM}), each model demonstrates unique strengths and weaknesses. For instance, although GPT-4o-mini achieves the best overall performance, it struggles with understanding updated information and is more sensitive to noise. In contrast, GPT-4o performs significantly better on these two aspects. When controlling for the base LLM and comparing retrievers (Figure \ref{fig:radar_retriever}), bge-m3 shows a well-rounded performance across all dimensions, explaining its overall superiority.

\section{Conclusion}
This paper makes two primary contributions: a synthetic data generation pipeline that produces realistic private user data, alongside a benchmark for evaluating how well AI models understand personal information from such data. 

Although many recent systems rely on RAG to produce personalized responses, our results suggest that these methods are currently oversimplified for real-world scenarios. In practice, user information is often noisy, fragmented, and interspersed with irrelevant data. Our findings highlight the need for more sophisticated methodologies and system designs that can effectively extract and leverage personal information from heterogeneous and imperfect private user data.



\section{Limitation}
\begin{itemize}
    \item Restricted Release of Ground-Truth Profiles: Although we plan to open-source the generated documents used for evaluation, we will not release the underlying “source of truth” profiles or the actual template used. This decision helps prevent potential misuse or cheating in downstream evaluations. However, we have provided a detailed description of our data generation pipeline to ensure transparency.
    \item Ethical Considerations: While every piece of information in the dataset is entirely synthetic and does not include any real personal data, and although we have taken steps to ensure realism and minimize harm, some content may still be perceived as offensive by certain people. We encourage users of this dataset to remain mindful of potential sensitivities and to apply further appropriate content filtering if necessary.
    \item Inconsistencies and Unrealistic Information: Given the large volume of automatically generated utterances, the synthetic documents may contain occasional inconsistencies or unrealistic details. Although we have implemented multiple checks and constraints, perfect coherence cannot be guaranteed. Future improvements to the generation process can further reduce these artifacts.
\end{itemize}

\section{*Ethical Statement}
PersonaBench is designed solely for evaluation purposes to assess AI models' ability to understand personalized information. The dataset is entirely synthetic and does not contain any real personal data. It is important to emphasize that the synthetic profiles and conversations do not reflect any real-world individuals, behaviors, or societal patterns. Any resemblance to real persons or events is purely coincidental.

This dataset is intended strictly for research use. It is not meant to be used for commercial applications or decision-making systems that could impact individuals or communities. We explicitly discourage any use cases that extend beyond the scope of academic research and evaluation.

While every effort has been made to ensure realism and minimize harm, we acknowledge that some generated content may be perceived as offensive or sensitive. We encourage users to remain mindful of potential sensitivities and to apply additional content filtering when necessary.

By using this dataset, researchers agree to uphold ethical standards, including responsible data handling and respect for societal norms.
\bibliography{custom}

\appendix

\section{Appendix}
\label{sec:appendix}
\subsection{User Profile Example}
\label{sec:profile_example}

Figure \ref{fig:profile} illustrates examples of user profiles for two characters. Each profile includes personal attributes categorized into three meta-categories: Demographic Information, Psychographic Information, and Social Information. The personal attributes are aligned with their respective social relationships.

\begin{figure*}[] 
\caption{User profile examples}
\label{fig:profile}
\begin{lstlisting}
"Profiles": [
    {
        "Name": Nicholas Torres,
        "Demographic Information":
            [
                Demographics.Gender.gender: Male,
                Demographics.Employment.Profession.profession: Historian,
                Demographics.Employment.Company.company_name: World History Society,
                Demographics.Location.Residence.country: United Kingdom,
                Demographics.Possession.Vehicle.vehicle, Volkswagen Golf,
                ...
            ],
        "Psychographic Information:
            [
                Psychographics.Preference.Book.Genre.book_genre: [Historical; Philosophy; Fantasy; Non-fiction; Classic Literature],
                Psychographics.Preference.Music.Instrument.music_instrument: [Piano; Violin; Guitar; Flute],
                Psychographics.Hobby.Activity.activity: [drawing, painting, bird watching],
                ...
            ],
        "Social Information":
            [
                Social_relationship.daughter: Sophia Torres,
                Social_relationship.colleague: Laura Schmidt,
                Social_relationship.friend: Samuel Green
            ],
    },
    {
        "Name": Sophia Torres,
        "Demographic Information":
            {
                Demographics.Gender.gender: Female,
                Demographics.Location.Birthplace.city_state: Oxford, England,
                Demographics.Employment.Job_Status.job_status: Student,
                Demographics.School.Degree.Subject.degree_subject: International Relations,
                ...
            }
        "Psychographic Information":
            {
                Psychographics.Preference.Movie.Title.movie_title: [The King's Speech, Hidden Figures, The Constant Gardener, Pride and Prejudice, Amelie],
                Psychographics.Preference.Sport.sport:
                [yoga, swimming, tennis, cycling, badminton],
                Psychographics.Want.Ability.ability: mastering multiple languages,
                ...
            }
        "Social Information":
            {
                Social_relationship.father: Nicholas Torres,
                Social_relationship.friend: Heidi Muller
            }
    }
    ...
]
\end{lstlisting}
\end{figure*}

\subsection{Social Graph Details}
\label{sec:social_graph_details}
Below are the short persona descriptions for each node in Figure \ref{fig:social}. The first three persona description are sampled from PersonaHub \cite{chan2024scaling}. The other personas are generated using GPT-4o.

\begin{itemize}
    \item Nicholas Torres: A historian specializing in the study of war crimes and their impact on reconciliation processes.
    \item Jennifer Moran: A Kenyan citizen who experienced police harassment.
    \item David Hess: A middle-aged man who lives in Germany, is fond of chess, and is passionate about sharing knowledge online.
    \item Laura Schmidt: An adventurous traveler and foodie who blogs about her experiences.
    \item Samuel Green: An easygoing librarian with a keen interest in rare books and archival material.
    \item Aisha Ouma: An energetic community organizer focused on youth empowerment in Nairobi.
    \item Heidi Muller: A sincere and hardworking nurse in a Munich hospital.
    \item Oliver Brown: A computer scientist working on AI ethics.
    \item Sophia Torres: A thoughtful university student studying international relations.
    \item Daniel Moran: An enthusiastic photographer known for his street photography.
\end{itemize}

\subsection{High-level Prompt Illustration for Dataset Generation}
\label{sec:prompt_example}
Figure \ref{fig:prompt_graph} illustrates the high-level concept of the prompt used for generating social graphs. Similarly, Figure \ref{fig:prompt_conversation} depicts the high-level concept of the prompt used for generating conversations.

\begin{figure*}[] 
\caption{Simplified representation of the prompt used to generate the social graph, highlighting the high-level structure and intent while omitting detailed instructions.}
\begin{lstlisting}
{
    "System": "You are an expert data generator tasked with creating realistic, synthetic social graphs with plausible social structures and connections. Each graph consists of nodes representing individuals and edges representing relationships. Follow these guidelines to ensure a coherent and realistic output: 
    /* Detailed graph generation guidelines */",
    
    "Instruction": "Given the following existing nodes: [{\"Name\": \"Persona\"}, {\"Name\": \"Persona\"}, ...], please expand the social graph to include more diverse people with distinct personas. 
    /* Detailed instruction guidelines */"
}

\end{lstlisting}
\label{fig:prompt_graph}
\end{figure*}

\begin{figure*}[] 
\caption{Simplified representation of the prompt used to generate conversation, highlighting the high-level structure and intent while omitting detailed instructions.}
\begin{lstlisting}
{
    "System": "You are an expert data generator specializing in creating realistic, multi-round conversations between two individuals: {person1} and {person2}. You will be provided with personal information about one of these users, such as their occupation, preferences, birthplace, or favorite artists. Your objective is to subtly integrate this information into the conversation in a way that feels natural and authentic. 
    /* Detailed conversation generation guidelines */",
    
    "Instruction": "Generate a realistic conversation between {person1} and {person2}. 
    /* Detailed instruction guidelines, including the sampled attributes and previous conversation history */"
}
\end{lstlisting}
\label{fig:prompt_conversation}
\end{figure*}


\subsection{Document Example}
\label{sec:document_example}
Figures \ref{fig:conversation_example}, \ref{fig:user_ai_interaction_example}, and \ref{fig:purchase_history_example} show examples of a single session for conversation, user-AI interaction, and purchase history documents, respectively.

\begin{figure*}[] 
\caption{An example conversation session between two socially connected individuals. The session reveals one person, Nicholas Torres’s attribute: Psychographics.Preference.Music.Artist.music\_artist = Hans Zimmer.}
\label{fig:conversation_example}
\begin{lstlisting}
"Conversations": [
    {
        "session": "conversation---session---1",
        "time": "2024/Oct/15/05:32 PM",
        "target_name": "Samuel Green",
        "conversation": [
            {
                "role": "Nicholas Torres",
                "content": "Hey Samuel, how's your day going so far?"
            },
            {
                "role": "Samuel Green",
                "content": "Hey Nicholas! It's going alright. Just another busy day at work. How about you?"
            },
            {
                "role": "Nicholas Torres",
                "content": "Not too bad, just finished up with some reports. Now I'm just relaxing with some music."
            },
            {
                "role": "Samuel Green",
                "content": "Nice. Anything interesting you've been listening to?"
            },
            {
                "role": "Nicholas Torres",
                "content": "Yeah, I've been diving into some film scores lately. There's something about those orchestral vibes that's really soothing after a long day."
            },
            {
                "role": "Samuel Green",
                "content": "Film scores are great! They can be really powerful. Do you have a favorite composer?"
            },
            {
                "role": "Nicholas Torres",
                "content": "Definitely. Hans Zimmer's work is incredible. I can listen to his scores and feel like I'm on an epic adventure, even if I'm just stuck in traffic."
            },
            {
                "role": "Samuel Green",
                "content": "Absolutely, his music is pretty iconic. I remember his score from 'Inception' was mind-blowing."
            },
            {
                "role": "Nicholas Torres",
                "content": "Inception is a masterpiece! It's amazing how music can transform a movie completely. Do you ever find yourself humming those tunes later in the day?"
            },
            {
                "role": "Samuel Green",
                "content": "All the time! Especially those intense tracks that just get stuck in your head. It's like having a mini soundtrack for your own life."
            },
            {
                "role": "Nicholas Torres",
                "content": "Exactly! It's like creating your own personal movie moments. So, any plans for the weekend?"
            },
            {
                "role": "Samuel Green",
                "content": "Thinking of hitting the mountains for a hike if the weather holds up. How about you? Any epic plans with some Zimmer tunes in the background?"
            },
            {
                "role": "Nicholas Torres",
                "content": "That sounds awesome. No major plans yet, but I might revisit some of his scores and let them guide an epic culinary adventure in the kitchen!"
            }
        ],
    }
\end{lstlisting}
\end{figure*}

\begin{figure*}[] 
\caption{An example where a user asks an AI assistant questions that reveal personal information, specifically: Demographics.School.Type.school\_type = Public University.}
\label{fig:user_ai_interaction_example}
\begin{lstlisting}
"User_AI_interaction": 
{
    "session": "user_ai_interactionn---session---2",
    "time": "2024/Oct/16/09:25 AM",
    "conversation": [
        {
            "role": "user",
            "content": "Hey, I've got a big project due soon, and I'm feeling a bit overwhelmed. Any tips on how to manage my time better?"
        },
        {
            "role": "assistant",
            "content": "Time management can be really challenging, especially with university projects looming. One strategy is to break down tasks into smaller steps and allocate specific time slots for each one."
        },
        {
            "role": "user",
            "content": "That sounds helpful. I attend a public university, and sometimes it feels like the workload is never-ending."
        },
        {
            "role": "assistant",
            "content": "Public universities can certainly have a diverse and demanding curriculum. Do you also use any digital tools to keep track of deadlines, like a planner app?"
        },
        {
            "role": "user",
            "content": "I usually jot things down in a notebook, but maybe I should switch to something digital. Any recommendations?"
        },
        {
            "role": "assistant",
            "content": "There are plenty of great apps out there. Notion and Trello are quite popular among students. They can help you organize tasks visually, which might be beneficial for managing that public university workload."
        },
        {
            "role": "user",
            "content": "I'll definitely check those out. Speaking of which, I need a break from this workload. Got any good movie recommendations?"
        },
        {
            "role": "assistant",
            "content": "Absolutely! Have you seen 'The Grand Budapest Hotel'? It's a beautifully styled film with a great story, perfect for unwinding."
        },
        {
            "role": "user",
            "content": "I haven't seen that yet. I'll add it to my list. Thanks for the chat, this was really helpful!"
        },
        {
            "role": "assistant",
            "content": "You're welcome! Good luck with your project, and enjoy the movie! Feel free to reach out anytime you need more tips or just a chat."
        }
    ],
}
\end{lstlisting}
\end{figure*}

\begin{figure*}[] 
\caption{An example of a character’s purchase history revealing personal information: Psychographics.Preference.Sport.sport = rowing.}
\label{fig:purchase_history_example}
\begin{lstlisting}
"Purchase history":
    {
        "session": "purchase_history---session---1",
        "time": "2024/Oct/24/01:27 PM",
        "purchase_history": [
            {
                "title": "Concept2 Model D Indoor Rowing Machine with PM5 Performance Monitor",
                "description": "The Concept2 Model D rowing machine makes it easy to achieve a full-body workout at home. This machine tracks your progress with a reliable and user-friendly performance monitor.",
                "brand": "Concept2",
                "categories": [
                    "Sports & Outdoors",
                    "Exercise & Fitness",
                    "Cardio Training"
                ]
            },
            {
                "title": "Craft Non-Padded Rowing Gloves",
                "description": "Designed specifically for rowing, these gloves by Craft offer protection and grip without compromising on comfort and flexibility.",
                "brand": "Craft",
                "categories": [
                    "Sports & Outdoors",
                    "Boating & Sailing",
                    "Rowing"
                ]
            },
            {
                "title": "Stamina BodyTrac Glider 1050 Rowing Machine ",
                "description": "A compact and portable rowing machine that brings a full-body, effective cardio workout to any home gym. It's ideal for smaller spaces.",
                "brand": "Stamina",
                "categories": [
                    "Sports & Outdoors",
                    "Exercise & Fitness",
                    "Cardio Training"
                ]
            }
        ],
    }
\end{lstlisting}
\end{figure*}


\end{document}